%% file: Paper-5388.tex
\documentclass[runningheads]{llncs}

\usepackage[utf8]{inputenc}
\usepackage[T1]{fontenc}
\usepackage{graphicx}
\usepackage{amsmath,amssymb}
\usepackage{booktabs}
\usepackage{multirow}
\usepackage{algorithm}
\usepackage{algpseudocode}
\usepackage{xcolor} 
\usepackage{subcaption}

\usepackage{pgfplots} 
\pgfplotsset{compat=1.18}
\newcommand{\datasetname}{Genodisc}

\begin{document}

\title{Be Indiscrete: The Benefits of Learning Continuous Spine Degeneration Severity Scores}

\titlerunning{Continuous Severity Ranking for Spine Pathology}

\author{%
Maria Monzon\inst{1,2}\and
Andrew Zisserman\inst{3} \and
Robin Y. Park\inst{3}\and
Catherine R. Jutzeler\inst{1,2}\and
Amir Jamaludin\inst{3}
}
\authorrunning{M. Monzon et al.}

\institute{%
 Biomedical Data Science Lab, Dept.  D-HEST, ETH Zurich, Zurich, Switzerland
\and Swiss Institute of Bioinformatics (SIB), Lausanne, 1015, Switzerland\\
\and
Visual Geometry Group, Dept.\ of Engineering Science,
University of Oxford, UK\\
\email{ mmonzon@ethz.ch}
}

\maketitle

\begin{abstract}
Lumbar spine degeneration is a major contributor to chronic low back pain and is routinely assessed on MRI using ordinal grading systems, e.g.\ normal, mild, moderate, severe. Consequently, most approaches to train models to grade these MRIs formulate grading as a multi-class classification problem, treating ordinal grades as categorical, ignoring differences in misclassification severity, and imposing hard decision boundaries on a continuous disease process. This work explores modeling spinal degeneration as a continuous severity ranking problem. We introduce SpineRankNet, a framework that learns scalar severity scores from lumbar spinal MRI, and compare it against multi-class classification and ordinal regression. 
Using multiple degeneration measures from the Genodisc dataset, 
we show that a model trained using a ranking loss
to produce a continuous score enables fine-grained ordering of MRI scans. Furthermore, the ordinal grading classes can be recovered from the score with comparable accuracy to those from a model trained directly for classification. 
The score learned by ranking even improves discrimination between more distant classes.
Source code is available at \url{https://github.com/spinetools/spineranknet}.

\end{abstract}


\section{Introduction}
Lumbar spine degeneration is a major cause of chronic low back
pain, expected to affect over 800 million people by 2050~\cite{lbp-GBD2021}.
MRI enables non-invasive assessment of intervertebral disc (IVD)
degeneration~\cite{Pfirrmann2001}, central canal and foraminal stenosis~\cite{Lee2011-Stenosis}, and vertebral endplate defects, among other conditions routinely graded on ordinal scales reflecting a continuous, progressive disease process.

Clinically, spinal degeneration is assessed using ordinal gradings: for example, Pfirrmann grading~\cite{Pfirrmann2001} assigns intervertebral discs (IVDs) to one of five categories reflecting progressive signal loss and disc height reduction on T2-weighted sagittal MRI, while central canal stenosis is graded on a four-point scale (``normal'', ``mild'', ``moderate'', ``severe'')~\cite{Lee2011-Stenosis}. Borderline cases often fall between adjacent categories, contributing to the low-to-moderate inter-rater agreement reported in spinal MRI assessment~\cite{Herzog2017}. 
Recent deep learning methods have achieved expert-level performance in automating radiological grading of spinal MRI~\cite{Jamaludin2017SpineNet,Windsor2024SpineNetv2,Windsor2024-SCT,Cina2023-ModicChanges}. However, most of these approaches are trained using categorical cross-entropy, treating ordinal grades as independent classes (which fundamentally mismatches the underlying biology) and penalising all misclassifications equally, despite the continuous nature of degeneration (e.g.\ ``severe'' misclassified as ``moderate'' vs ``normal''). Moreover, IVD degeneration is a continuous biological process, and collapsing it into discrete labels discards nuance, potentially grouping biologically distinct disease states under a single label. Many spinal MRI pathologies also exhibit extreme class imbalance, where severe cases are rare, leading standard classifiers to bias toward the majority class and underperform on clinically critical minority cases. Simplifying the problem to binary classification (normal vs.\ abnormal) as in~\cite{Jamaludin2017SpineNet} also sacrifices clinically meaningful granularity. 

This work models spinal degeneration as a continuous value, which overcomes the problems identified above with categorical scoring. First, we reformulate training as a ranking problem, learning a continuous scoring function that preserves the relative ordering of disease severity, instead of predicting absolute labels. Penalisation for error in ranking during training preserves the severity gap, making large errors (e.g.\ ``normal” vs ``severe'') explicitly more costly than adjacent misclassifications (e.g.\ ``mild” vs ``moderate”).  The resulting score, learned by ranking, also captures the within-grade variability and naturally mitigates class imbalance, since severe cases appear in many training pairs regardless of their absolute frequency.  Despite these advantages, ranking has been largely unexplored in structured ordinal grading of spinal pathology.

A continuous severity score is also more practical than discrete bins. It can serve as a more sensitive measure for detecting subtle longitudinal changes, supporting earlier identification of disease progression or treatment response. It would also facilitate relative comparisons during annotation, allowing clinicians to characterise the nuance of disease, rather than forcing an absolute categorical decision. By preserving ordinal structure on a continuous scale, the framework allows for more clinically-consistent transitions across grades that better reflect the gradual nature of spinal degeneration.

In summary, we make the following contributions:  (\emph{i}) we introduce \textit{SpineRankNet}, a multi-pathology framework that produces continuous severity scores for lumbar spine MRI grading; (\emph{ii}) we conduct a large-scale comparison of classification, ordinal regression, and ranking across 11 concurrent spinal grading tasks on the multicentre Genodisc cohort;  (\emph{iii}) we show that ranking consistently reduces large-grade errors; and (\emph{iv}) we demonstrate that a more granular, continuous severity scale allows for better separation of borderline cases.


\subsection{Related Work}

Radiological assessment of lumbar degeneration has traditionally relied on geometric biomarkers, i.e.\ disc height ratios, canal diameters, and foraminal widths derived from manual or semi-automated measurements~\cite{Pfirrmann2001,Modic1988}.
Recent models use  either segmentation-derived metrics~\cite{Neubert2013,Natalia2020,Zheng2022-BianqueNet} or end-to-end deep learning and on  Pfirrmann grading~\cite{Kowlagi2023-PfirmannBaseline,Minin2025-SpineScan}, stenosis~\cite{Lu2018-DeepSpine}, herniation~\cite{Lewandrowski2020-Herniation}, and Modic changes~\cite{Cina2023-ModicChanges}, but  most existing models have focused only on individual pathologies.
Among these,  SpineNet~\cite{Jamaludin2017SpineNet,Windsor2024SpineNetv2} and the Context-Aware Spinal Transformer~\cite{Windsor2024-CAST,Windsor2024-SCT} are notable exceptions, performing concurrent multi-pathology grading 
within a unified architecture.
However, they still treat ordinal severity labels as independent nominal classes and ignore the ordering and magnitude of grading errors.

Recent ordinal regression methods~\cite{Pedregosa2014-OrdinalRegression,Shi2023-CORN} have partially addressed this issue by enforcing rank consistency, demonstrating improvements over nominal cross-entropy in medical grading tasks \cite{Saffari2015}. However, they still discretise the severity continuum using fixed thresholds, leading to unstable decisions at grade boundaries and collapsing within-grade variability into a single label.
Pairwise ranking offers an alternative by learning a continuous scoring function from relative severity comparisons rather than categorical labels~\cite{Hullermeier2008-PairwiseRanking}. Such relative-attribute frameworks are well established in computer vision~\cite{Parikh2011-RelativeAttributes,Souri2016-DeepRelativeAttributes,Burges2005-RankNet,Ahmed2021-DeepRankSVM}, where continuous scores from relative comparisons are highly discriminative for ordered attributes.
In medical image analysis, ranking has been applied to preserve
clinical severity ordering from imaging data~\cite{Pedregosa2012-LearningRank},  to learn continuous scores for single conditions such as knee osteoarthritis~\cite{Li2020}. 
However, these applications remain limited to individual pathologies and have not been systematically compared against classification and ordinal regression baselines. 
To our knowledge, pairwise ranking has never been applied to structured multi-pathology severity grading of the spine.

\section{Spine Degeneration Continuous Score}

\begin{figure}[tbp]
    \centering

    \includegraphics[width=0.95\linewidth]{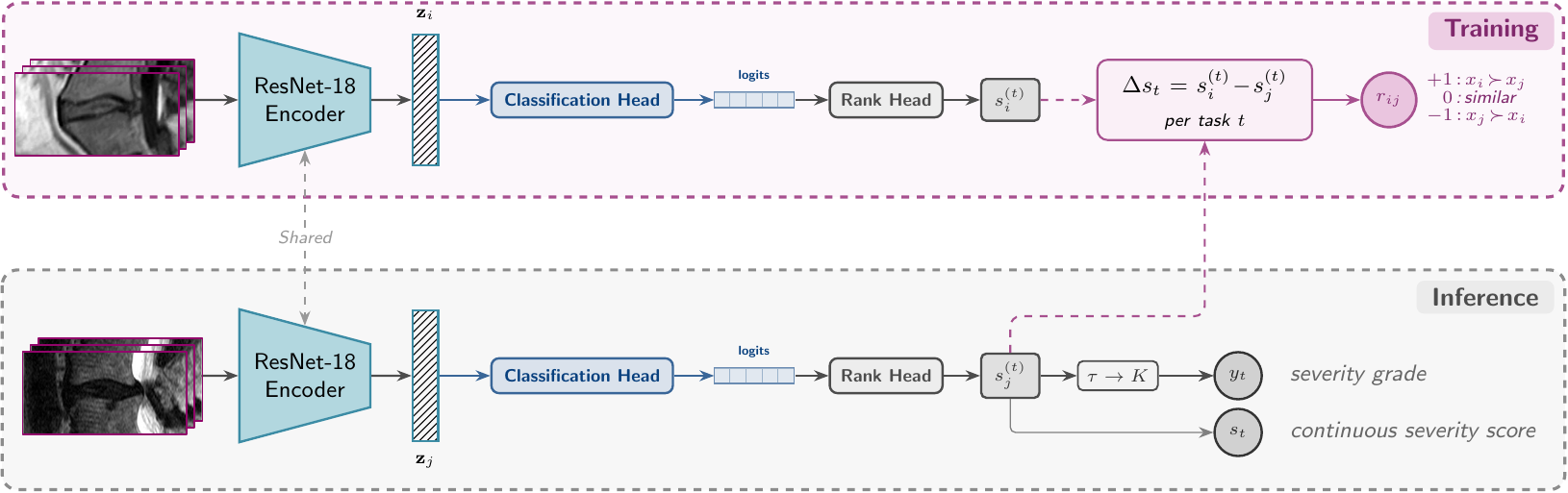}
    \caption{\textbf{Multi-task architecture for joint classification and ranking.} A shared 3D ResNet-18 encoder produces an embedding $\mathbf{z}$, followed by task-specific classification and ranking heads. Each ranking head outputs a continuous severity score $s^{(t)} \in [0,10]$; for clarity, the figure shows one representative task head, though there are heads for each of the 11 spinal grading tasks. Pairwise supervision is derived from ground-truth grade orderings and optimized with a severity-aware hinge loss.}
\label{fig:architecture}
\end{figure}
\label{sec:ranking}

This section describes our method to predict a continuous score for each grading task from a single IVD input. Unlike classification, which treats grades as categorical, or ordinal regression, which discretises at fixed thresholds, our ranking formulation preserves the continuous disease spectrum while learning from relative severity comparisons. At inference, our model outputs both a continuous score and a corresponding radiological grade (see Fig.~\ref{fig:architecture}). We train this model using pairwise ranking procedure, which is described next.

\subsection{Ranking to be indiscrete}

For each grading task (e.g.\  Pfirrmann or narrowing) our aim is to learn a scalar scoring function for the IVD $x$:
\[
s_t^{(\theta)} : x \to [0,S],
\]
such that the score is aligned with the $K$ ordinal
clinical grades for that task $y_t \in \{0,\ldots,K_{t-1} \}$. 
We set $S{=}10$ across all tasks, as this range aligns with common clinical degeneration scales.

We use pairwise ranking supervision to learn the scoring function, and extend the ranking SVM loss~\cite{Parikh2011-RelativeAttributes,Ahmed2021-DeepRankSVM} with severity-aware components to achieve this.



\paragraph{\textbf{{Ranking loss.}}}
To train the model on pairwise comparisons, we construct pairs $(x_i, x_j)$ with relative labels $r_{ij}^t \in \{-1, 0, +1\}$ indicating whether IVD $x_i$ is less severe than, equally severe to, or more severe than disc $x_j$ for task $t$.


A severity-weighted hinge loss is applied to ordinal pairs  $\mathcal{P}^t = {(x_i,x_j) }$  with different grades (${ y_i^t \neq y_j^t}$):
$$\mathcal{L}_{\text{hinge},m}^t = \frac{1}{|\mathcal{P}^t|} \sum_{(x_i,x_j)\in\mathcal{P}^t}  w_{ij}^t \left[\ \max(0, m_{ij}^t - r_{ij}^t \Delta s_{ij}^t)\right]^2 \quad  \quad  \text{if } d_{ij}^t \neq 0$$
where  $r_{ij}^t = \text{sign}(y_i^t - y_j^t) \in {-1, +1}$ is the label direction, and $\Delta s_{ij}^t = s_i^t - s_j^t$ is the score difference. This term penalizes both ranking errors (wrong direction) and violations of the margin requirement. The severity-aware margin, $m_{ij}^t$, and weight, $w_{ij}^t$ are defined below.

To prevent score spread within a single grade from fragmenting the learned
severity axis, same-grade pairs are penalised by a squared score difference \cite{Parikh2011-RelativeAttributes}:
\begin{equation}
  \mathcal{L}_{\text{sim}}^t =
    \frac{1}{|\mathcal{S}^t|}
    \sum_{(x_i,x_j)\in\mathcal{S}^t}
    \bigl(\Delta s_{ij}^t\bigr)^{2} \quad  \quad  \text{if } d_{ij}^t = 0
  \label{eq:sim}
\end{equation}

The total training loss sums over all $T$ tasks combining the two terms for each task $t$:
\begin{equation}
  \mathcal{L} = \sum_{t=1}^{T} \mathcal{L}_{\mathrm{rank}}^t
  = \sum_{t=1}^{T} \left( \lambda_{dis}\mathcal{L}_{\text{hinge}_{m}}^t +\lambda_{sim}\mathcal{L}_{\text{sim}}^t \right)
\end{equation} 
where $\lambda_{dis} = 1 $, $\lambda_{sim} = 0.25$ represents the trade-off coefficients between enlarging the margin and satisfying the pairwise relative constraints \cite{Ahmed2021-DeepRankSVM}.

\paragraph{\textbf{{Difference-specific weight.}}}
To reflect the clinical significance of grade differences, each ordinal pair is weighted by its normalized grade distance:
\begin{equation}
w_{ij}^t = |y_i^t - y_j^t| \cdot \frac{S}{K_t - 1},
\label{eq:weight}
\end{equation}
where $y_i^t, y_j^t$ are ground-truth gradings, and the denominator $K_t - 1$ normalizes weights across tasks with different grade granularities.

\paragraph{\textbf{{Severity-aware margin.}}}
Unlike standard ranking SVM~\cite{Parikh2011-RelativeAttributes,Ahmed2021-DeepRankSVM} that applies a fixed margin $m$ to all pairs, we  adapt  the margin to both the severity gap and the task-specific grade structure $$m_{ij}^t = m_{\text{base}} + m_{\text{scale}} \cdot \frac{w_{ij}^t}{S}$$ with $m_{\text{base}} = 0.5$ and $m_{\text{scale}} = 1.5$, adjacent grades (small $w_{ij}$) require minimal margin, while extreme pairs (large $w_{ij}$) or severe misrankings require proportionally larger score separations (to overcome the poor scaling critiqued in~\cite{Souri2016-DeepRelativeAttributes}).

\paragraph{\textbf{Score Parameterization}}

For a given input IVD volume $x^{(i)}$,  we first extract a shared feature
representation using a convolutional encoder. 
\begin{equation}
\mathbf{z}^{(i)} = f_\theta(x^{(i)}) \in \mathbb{R}^d,
\end{equation}
shared across all $T$ grading tasks. Per-task linear heads first map $\mathbf{z}^{(i)}$ to grade logits $\hat{y}_t^{(i)} \in \mathbb{R}^{K_t}$ ($K_t$ ordinal grades), pretrained with inverse-frequency-weighted cross-entropy; the ranking heads are then trained on top to produce continuous severity scores
(Fig.~\ref{fig:architecture}). For each task $t$, we learn a lightweight
projection head
\[
\varphi_t : \mathbb{R}^d \to \mathbb{R},
\]
implemented as a small two-layer MLP, that maps the shared embedding to
a scalar latent score
\begin{equation}
h_t^{(i)} = \varphi_t\!\left(\mathbf{z}^{(i)}\right).
\end{equation}

The latent score is then transformed into a bounded continuous severity
score
\begin{equation}
s_t^{(i)} = \sigma_S\!\left(h_t^{(i)}\right)
= S \cdot
\frac{\mathrm{sp}\!\bigl(\beta\, h_t^{(i)}\bigr)}
     {\mathrm{sp}\!\bigl(\beta\, h_t^{(i)}\bigr)
      + \mathrm{sp}\!\bigl(\beta\,(S - h_t^{(i)})\bigr)},
\label{eq:score}
\end{equation}
where $\mathrm{sp}(z) = \ln(1 + e^{z})$ is the softplus function and
$\beta = 5$ controls the boundary sharpness. This soft-clamp guarantees
$s_t^{(i)} \in [0, S]$.

To obtain a discrete severity grade from the continuous severity score, we introduce $K{-}1$ ordered thresholds $\boldsymbol{\tau} = \{\tau_1, \dots, \tau_{K-1}\}$ that partition each task-specific score into $K$ ordinal categories.

\section{Dataset and Implementation Details}
\textbf{Dataset.}
The Genodisc dataset~\cite{Jamaludin2017SpineNet} includes T1w and T2w sagittal lumbar MRIs from ${\sim}2{,}000$ subjects across five European centres with heterogeneous acquisition parameters (0.2--3.0\,T, 0.25--0.9\,mm in-plane, 2.6--6.0\,mm slice spacing).
Each subject has up to six lumbar IVD levels (T12-L1 to L5-S1). 
Expert radiological annotations cover 11 grading tasks per IVD level, spanning disc degeneration (Pfirrmann grading, disc narrowing, herniation, bulging), spinal
canal and nerve compression (central canal stenosis, foraminal stenosis), structural abnormalities and vertebral degeneration (spondylolisthesis, endplate changes).  The dataset exhibits severe class imbalance: 60-96\% normal prevalence, with severe cases comprising only 2--10\% (e.g., severe stenosis: 2\%). 
 
IVD volumes ($128 \times 256 \times 12$ voxels) are extracted via SpineNetv2~\cite{Windsor2024SpineNetv2} vertebra localization, disc-centered rotation, and resampling.
Data is split at the patient level into train ($80\%$), validation ($10\%$), and test ($10\%$),
stratified by acquisition centre and pathological grade distribution 
to ensure representative class prevalence across all partitions.


\noindent \textbf{Evaluation.}
We report balanced accuracy (BA), quadratic weighted Cohen’s kappa (QWK), receiver operating characteristic area under the curve (ROC-AUC), Matthews correlation coefficient (MCC) and mean absolute error (MAE) on a held-out test set.  
Deterministic test-time augmentation (TTA) evaluates each scan across 54 views (2 flips × 3 horizontal translations × 3 vertical translations × 3 slice shifts). Final predictions average the softmax probabilities or predicted severity scores across views.

\noindent \textbf{Architecture and training.}
All implementations share a 3D ResNet-18 encoder with asymmetric convolutions
(strides $(1,s,s)$) that preserve the slice dimension, reflecting the
anisotropic resolution of spinal MRI.
For each task~$t$, a linear head maps the pooled features
$\mathbf{z} {=} \mathrm{GAP}(\Phi(\mathbf{x})) \!\in\! \mathbb{R}^{512}$
to logits $\hat{\mathbf{y}}_t \!\in\! \mathbb{R}^{K_t}$, trained with
inverse-frequency-weighted cross-entropy
($w_c {=} N_t / (K_t {\cdot} N_{t,c})$).
T1w/T2w sequences are randomly sampled during training, with Pfirrmann/disc narrowing labels masked for T1w. Encoder and classification heads are trained from scratch (AdamW, lr=$10^{-3}$, wd=$10^{-4}$) for up to 300 epochs (early stopping: patience 30, min 50 epochs; batch size 32). Ranking heads are then optimized for 200 epochs while fine-tuning the encoder (lr=$10^{-4}$) to prevent catastrophic forgetting. Thresholds $\boldsymbol{\tau}$ are grid-searched on validation data.

Training augmentations include sagittal slice shifts ($\pm 2$ slices,
with boundary slices zeroed), left-right slice-order flips
($p{=}0.5$), intensity offset ($\pm 0.1$), random translations (up to
32\,px horizontally, 24\,px vertically), isotropic scaling
($\pm 10$\,\%), and in-plane rotation ($\pm 15^{\circ}$).

 \noindent{\textbf{Pair construction for training.}} To construct pairs of training examples, we sample candidate pairs from $B$ samples in a mini-batch. 
For a given pair $(i,j)$, we calculate the signed difference between the grades, $d_{ij}^t = y_t^{(i)} - y_t^{(j)}$, for each grading task $t$. From the
$\binom{B}{2}$ candidates in a batch of $B{=}32$, 70\% of pairs per task are selected by descending violation magnitude (hard-negative mining) and 30\% sampled uniformly to preserve training stability.

\noindent \textbf{Classification, regression, and ranking.}
We benchmark against representative methods from three objective families under identical 3D ResNet-18 encoders and evaluation protocols: (i)~Classification (inverse-frequency-weighted CE); (ii)~Ordinal regression (CORN~\cite{Shi2023-CORN} with cumulative binary decomposition, direct MAE$_\ell$ regression); (iii)~Pairwise ranking (RankNet~\cite{Burges2005-RankNet}, DeepRankSVM~\cite{Ahmed2021-DeepRankSVM}). All ranking methods use two-phase training: classification heads pretrained with weighted CE, then ranking heads are finetuned.

\section{Experimental Results}

Our experiments aim to establish that ranking-trained continuous scores match direct classification performance when discretized, and to highlight their added clinical value over discrete ordinal grades.

 
\input{results/table-clf-loss}



\paragraph{Classification and ordinal baselines vs pairwise ranking.}
Table~\ref{tab:ordinal_loss_comparison} shows that classification (CE) and ordinal regression (CORN) impose a trade-off between accuracy and ordinal concordance, whereas pairwise ranking methods yield the best ordinal metrics. SpineRankNet achieves superior ordinal agreement (QWK 0.76, MAE 0.14) while maintaining classification-level discrimination (ROC-AUC 0.94). Notably, its lowest MAE demonstrates systematic reduction of large-grade prediction errors—critical for clinical safety—while highest MCC confirms robustness under severe class imbalance, all without compromising overall accuracy. It further outperforms prior ranking models (RankNet, DeepRankSVM), validating the severity-adaptive margin and same-grade similarity loss.


\paragraph{Ranking Layer Selection and Thresholding.}
For threshold selection, coordinate-descent grid search (coarse 200-point sweep, two refinements, Nelder--Mead polish) achieves BA 63.3\%/QWK 0.762, surpassing isotonic regression fitting monotonic score$\to$grade mappings (55.7\%/0.744), Youden's index optimizing per-boundary ROC thresholds (58.8\%/0.685), and GMM clustering scores into $K$ Gaussians (54.3\%/0.448). For the ranking head, the same two-layer MLP outperforms linear projection (58.3\%/0.733), Kolmogorov-Arnold Networks (61.0\%/0.698), and transformer heads (41.8\%/0.303).

\begin{figure}[tbp]
    \centering
    \includegraphics[width=0.95\linewidth]{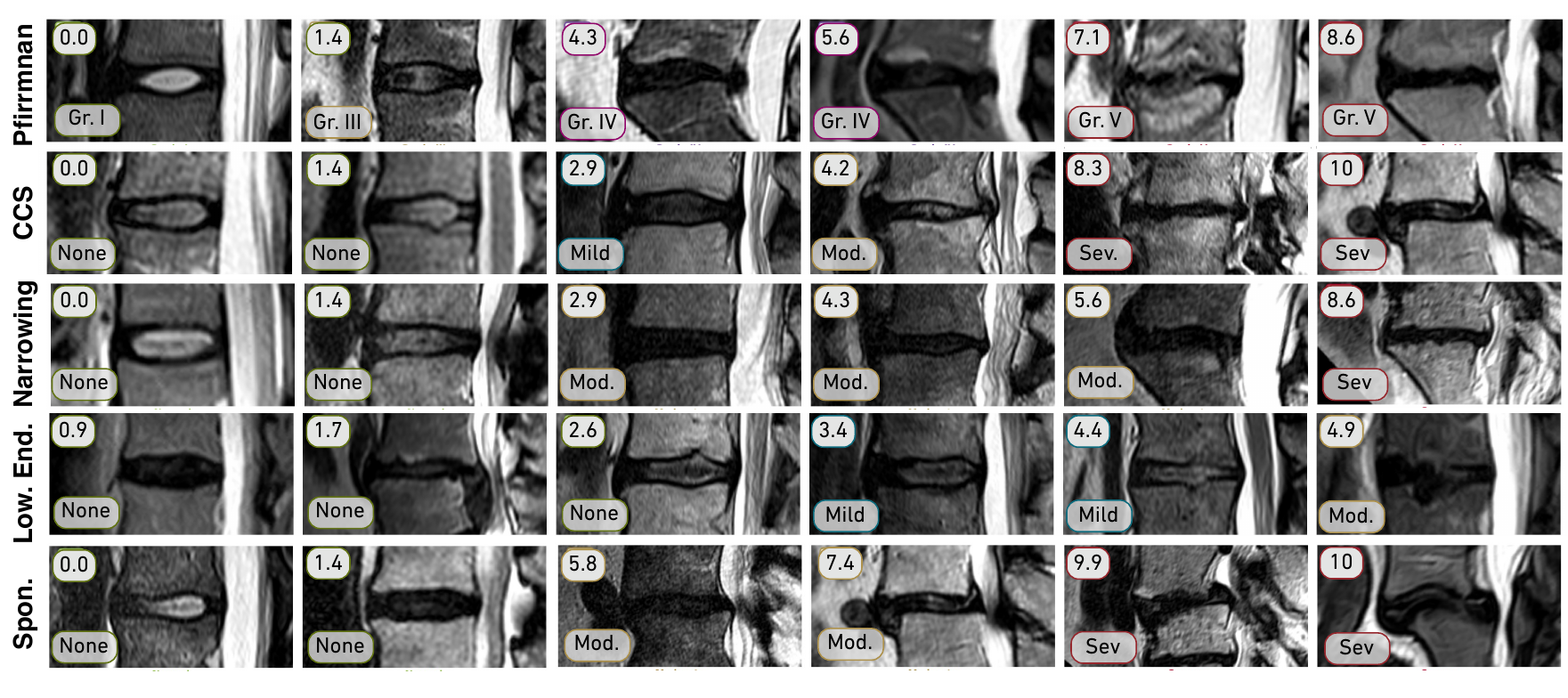}
        \caption{Example visualization of severity ranking for the pathology scores.
        IVD volumes for Pfirrmann, central canal stenosis (CCS), Narrowing, and Lower Endplate defects (Low.\ End) and Spondylolisthesis, ordered by predicted score $s \in [0,10]$ (top) with ground-truth ordinal grades (bottom). Note how the severity progressively increases with score. For example, 
        for CCS, \textit{``None''} cases (0.0 vs.\ 1.4) differentiate normal from subtle central canal compression, while ``Severe'' cases (8.3$\to$10) capture progressive cord impingement}

    \label{fig:ranking-examples}
\end{figure}

\paragraph{Continuous score degeneration.}
Fig.~\ref{fig:ranking-examples} further shows that the learned continuous severity scores capture intra-grade variability that categorical models cannot represent (e.g., mild-to-moderate discs span 2.7–4.1 despite identical grade labels; advanced degeneration cases cluster between 7.5–9.8) 
\noindent 
, consistent with the quantitative ordinal improvements 
reported in Table~\ref{tab:ordinal_loss_comparison}. 

\section{Conclusion and Future Work}
\label{sec:conclusion}

We provide a  comparison of classification, ordinal
regression, and pairwise ranking objectives for highly-imbalanced, multi-pathology spinal grading on lumbar MRI.
Experiments on the multi-centre Genodisc cohort show that ranking-based models shift the error profile toward adjacent-grade confusions while reducing clinically distant misclassifications, reflected by consistent QWK improvements across the majority of 11~grading tasks  and pairwise formulations.
The continuous severity scores also identify borderline samples near inter-grade thresholds, offering a finer-grained view of disease progression than discrete labels alone. 

While pairwise ranking shows promise for ordinal grading tasks where error magnitude matters more than per-class accuracy, all experiments use a single dataset. Future work should include 
external validation on truly independent cohorts, evaluation under longitudinal scenarios, and
assessment of clinical utility by radiologists. 

\subsubsection*{Acknowledgements.}
This project was supported by a grant (\#380, Jutzeler, Manjaly) from the
Strategic Focus Area ``Personalized Health and Related Technologies (PHRT)''
of the ETH Domain (Swiss Federal Institute of Technology), by the Swiss
National Science Foundation (Ambizione Grant \#PZ00P3\_186101, CRJ). This project was also funded by the EPSRC programme grant Visual AI (EP/T025872/1) and EPSRC CDT in Health Data Science
(EP/S02428X/1). We thank our clinical collaborators Prof.\ J.\ Fairbank, Dr.\ J.\ Urban, Dr.\ S.\ Ather, and Prof.\ I.\ McCall.
AI-assisted tools (GitHub Copilot, Claude, Perplexity, Writefull) were used for coding and grammar refinement; the authors reviewed all output and take full responsibility for the content.

\subsubsection*{Disclosure of Interests.}
Catherine R. Jutzeler serves as a scientific consultant to AbbVie. This role is unrelated to the present work and had no influence on the design, analysis, interpretation, or reporting of the study. 
The remaining authors have no relevant competing interests to declare.
\bibliographystyle{splncs04}
\bibliography{references}


\end{document}

%% file: results/table-clf-loss.tex
\begin{table}[t]
\centering
\caption{Comparison on the \datasetname{} test set (ResNet-18),  aggregated over 11 grading tasks.  
Mean\,$\pm$\,std over the individual task performances (single run)}
\label{tab:ordinal_loss_comparison}
{\fontsize{8}{9.5}\selectfont
\setlength{\tabcolsep}{4pt}
\renewcommand{\arraystretch}{0.95}
\begin{tabular}{@{}llccccc@{}}
\toprule
 & Loss
  & BA\,(\%)\,$\uparrow$
  & ROC-AUC\,$\uparrow$
  & QWK\,$\uparrow$
  & MAE\,$\downarrow$
  & MCC\,$\uparrow$ \\
\midrule
  & CE
    & $\mathbf{64.1 \pm 7.9}$
    & $\mathbf{0.94 \pm 0.02}$
    & $0.69 \pm 0.12$
    & $0.18 \pm 0.09$
    & $0.54 \pm 0.08$ \\
\midrule
  & CORN
    & $56.9 \pm 9.8$
    & ---*
    & $0.72 \pm 0.12$
    & $0.15 \pm 0.10$
    & $0.55 \pm 0.10$ \\
  & MAE$_\ell$
    & $58.8 \pm 15.7$
    & $0.84 \pm 0.13$
    & $0.54 \pm 0.24$
    & $0.39 \pm 0.30$
    & $0.38 \pm 0.20$ \\
\midrule
  & RankNet
    & $58.7 \pm 8.3$
    & $0.90 \pm 0.03$
    & $0.73 \pm 0.11$
    & $0.16 \pm 0.09$
    & $0.57 \pm 0.07$ \\
  & DeepRankSVM
    & $61.0 \pm 8.4$
    & $0.88 \pm 0.05$
    & $0.74 \pm 0.11$
    & $0.16 \pm 0.09$
    & $0.57 \pm 0.07$ \\
  & SpineRankNet
    & $63.3 \pm 9.5$
    & $\mathbf{0.94 \pm 0.04}$
    & $\mathbf{0.76 \pm 0.10}$
    & $\mathbf{0.14 \pm 0.09}$
    & $\mathbf{0.59 \pm 0.08}$ \\
\bottomrule
\end{tabular}}
\smallskip
\par\noindent{\tiny
*ROC for CORN not available}
\end{table}